\documentclass{article}

\usepackage[T1]{fontenc}
\usepackage{spconf,amsmath,amssymb,graphicx,hyperref}
\usepackage{booktabs}
\usepackage{array}

\newcommand{\tablebodyfont}{\fontsize{9pt}{10.5pt}\selectfont}

\hypersetup{hidelinks}

\title{SPHERICAL GEOMETRY-AWARE PANOPTIC SCENE GRAPH GENERATION FOR EQUIRECTANGULAR PANORAMAS}

\name{Zhe Wu$^{1}$, Xiaoyan Fei$^{2}$, Xinyi Jiang$^{1}$,
Yanxu Wei$^{1}$, Junming Rao$^{1}$, Yihong Yan$^{1}$,
Xuewei Li$^{1}$\sthanks{Corresponding author: xueweili@sdju.edu.cn.}}

\address{
$^{1}$Shanghai DianJi University, Shanghai, China\\
$^{2}$University of Minnesota Twin Cities, Minneapolis, MN, USA
}

\begin{document}

\maketitle

\begin{abstract}
Panoptic scene graph generation (PSGG) represents scene entities with
segmentation masks and directed relations. Equirectangular projection (ERP)
panoramas challenge PSGG through distortion, latitude-dependent pixel areas,
and an artificial seam. We propose a two-stage spherical geometry-aware
framework that fuses overlapping perspective-view predictions using
observation reliability, spherical overlap, and angular proximity. Relation
reasoning combines spherical-area-aware aggregation, weighted no-relation
supervision, and dual-longitude score fusion. We also construct a dataset of
340 high-resolution panoramas with 10,571 nodes and 6,632 directed relations.
Against corresponding baselines, spherical fusion improves PQ by 36.03\%
under matched views, while relation reasoning improves mR@50 by 11.20\%
under PredCls with ground-truth masks and object labels. The full framework
improves mR@50 by 52.99\% under SGDet with predicted panoptic segments.
These results demonstrate the value of spherical geometry for panoramic PSGG.
\end{abstract}


\section{Introduction}
\label{sec:intro}

Panoptic scene graph generation (PSGG) represents scene entities with
segmentation masks and directed semantic relations~\cite{Yang2022PSG}.
Panoramas capture full-scene entities and interactions in one observation.
Prior work exploits panoramic context for holistic 3D
understanding~\cite{Zhang2021DeepPano,Dong2024PanoContext} and studies
omni-range context~\cite{Yang2021Omni}, projection
distortion~\cite{Zhang2022Trans4PASS,Zheng2023DPPASS}, and spherical
geometry~\cite{Li2023SGAT4PASS,Li2026RELSF4PASS} in panoramic segmentation. PSGG models address
predicate imbalance~\cite{Zhou2023HiLo}, pair selection~\cite{Wang2024PairNet},
and mask-aware relation modeling and ranking through
DSFormer~\cite{Lorenz2024DSFormer}, but largely assume perspective images.
Panoramic PSGG must therefore model spherical geometry in both entity
representation and relation prediction.

\begin{figure}[!t]
    \centering
    \includegraphics[width=0.95\columnwidth]{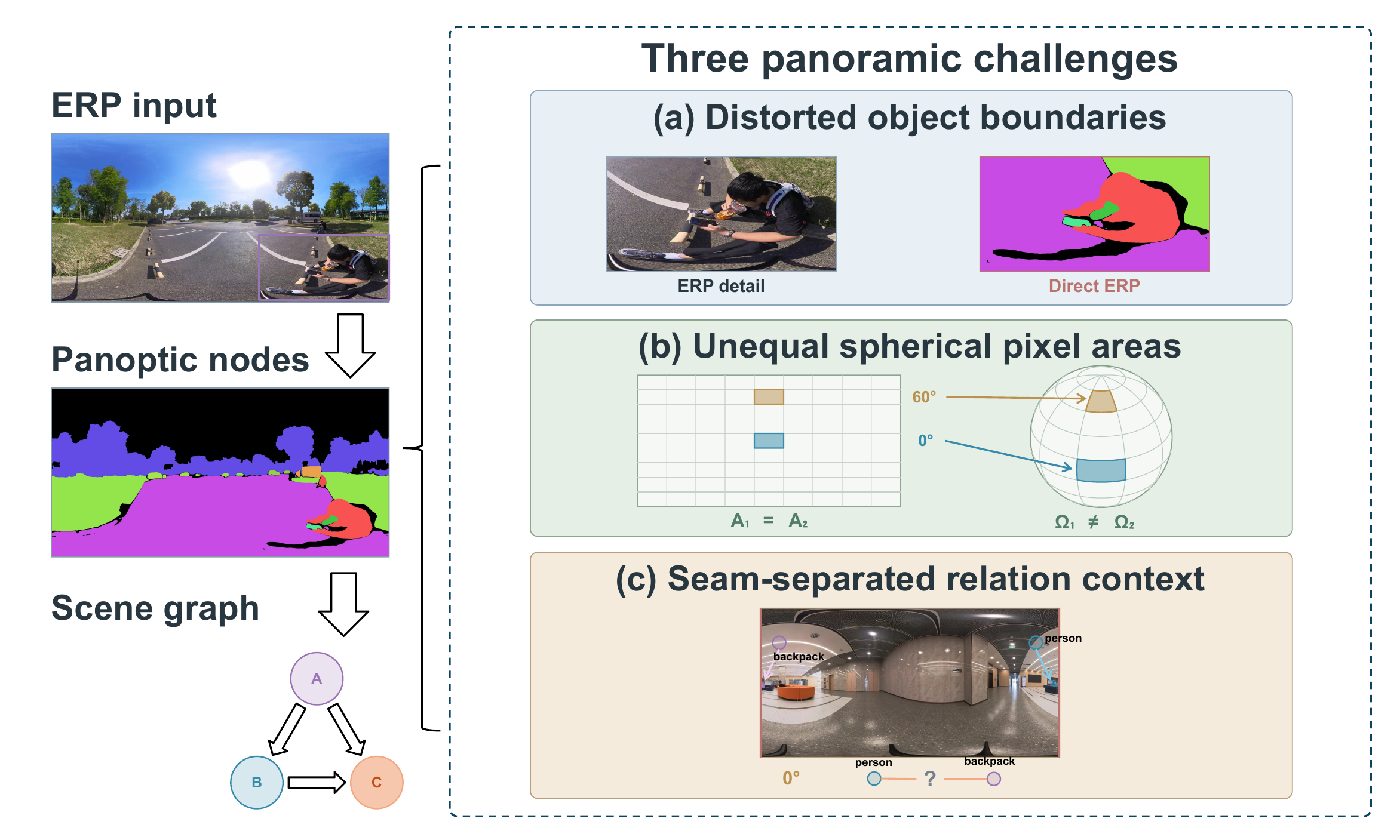}
    \caption{
    \textbf{Panoramic geometry challenges for PSGG.}
    (a) Direct parsing of ERP images introduces object boundary errors
    in the highlighted region.
    (b) Equally sized ERP regions correspond to different spherical
    surface areas at different latitudes.
    (c) The ERP seam artificially separates spatially adjacent instances
    on the sphere, disrupting their shared relational context.
    }
    \label{fig:geometry}
\end{figure}

Equirectangular projection creates three geometric challenges for
panoramic PSGG (Fig.~\ref{fig:geometry}): \textbf{distorted object
boundaries}, \textbf{unequal spherical areas for equal-sized image regions},
and \textbf{seam-separated relational context}. Overlapping perspective views
reduce distortion but yield duplicate or truncated instances with inconsistent
identities, requiring reliable association and fusion. Relation features should
reflect spherical rather than image area, and relation-existence predictions
should remain stable when the longitude origin moves the ERP seam without
changing the scene.

We make three contributions. (1) \textbf{Spherical geometry-aware panoptic
segmentation (SGPS)} associates and fuses cross-view instances using
observation reliability, spherical overlap, and angular proximity, improving
PQ by 36.03\% over a baseline with the same view configuration.
(2) \textbf{Spherical geometry-aware relation reasoning (SGRR)} combines
spherical-area-aware aggregation, weighted no-relation supervision, and
dual-longitude score fusion. Under PredCls, this module improves mR@50 by
11.20\% over its corresponding baseline. (3) \textbf{A high-resolution
panoramic PSGG dataset} comprises 340 panoramas, 10,571 nodes, and 6,632
directed relations with multi-stage quality control. Under SGDet, the complete
framework improves mR@50 by 52.99\% over the corresponding baseline, while
stage-wise ablation isolates the gains from SGPS and SGRR.

\section{Method}
\label{sec:method}

\begin{figure*}[!t]
    \centering
    \includegraphics[width=0.95\textwidth]{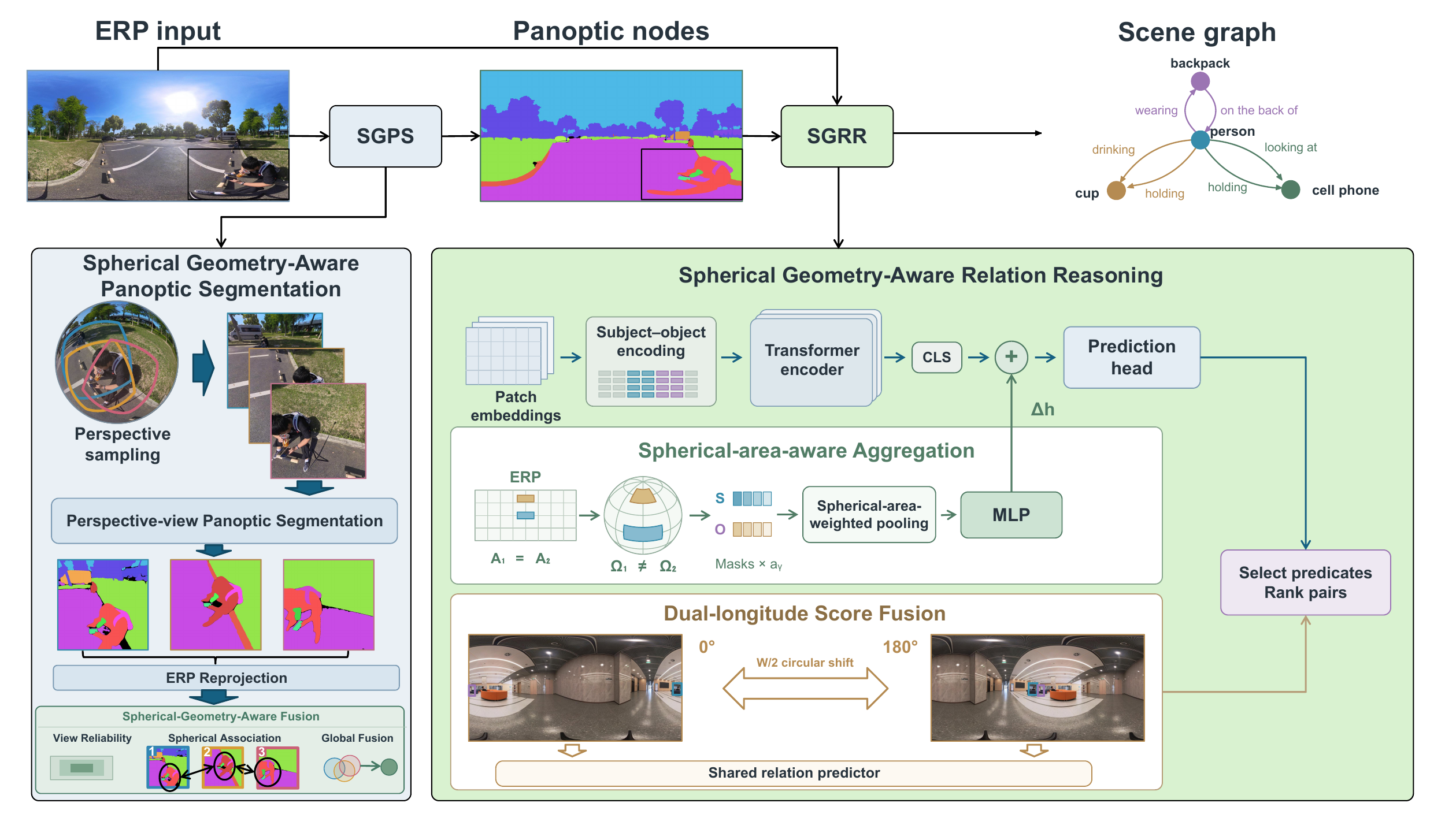}
    \caption{
    Spherical geometry-aware PSGG framework.
    Spherical geometry-aware panoptic segmentation (SGPS) fuses overlapping
    perspective-view predictions. Spherical geometry-aware relation reasoning
    (SGRR) combines spherical-area-aware aggregation, weighted no-relation
    supervision, and dual-longitude score fusion.
    }
    \label{fig:framework}
\end{figure*}

Panoramic PSGG maps an RGB ERP image
$I\in\mathbb{R}^{H\times W\times3}$ to a directed graph
$\mathcal{G}=(\mathcal{V},\mathcal{E})$, where $H$ and $W$ are the image
height and width. Each node $v_i=(M_i,c_i,b_i)\in\mathcal{V}$ contains a
\emph{thing} or \emph{stuff} mask $M_i$, object label $c_i$, and bounding box
$b_i$. Each directed edge assigns a semantic predicate to an ordered
subject--object pair. Spherical geometry-aware panoptic segmentation (SGPS)
constructs panoramic nodes, and spherical geometry-aware relation reasoning
(SGRR) predicts their directed relations. Figure~\ref{fig:framework} shows
the framework.

\subsection{Spherical Geometry-Aware Panoptic Segmentation}
\label{ssec:fusion}

A fixed MaskDINO~\cite{Li2023MaskDINO} predicts panoptic masks in 30
overlapping $90^\circ$ views sampled at yaw angles $60^\circ k$,
$k=0,\ldots,5$, and pitch angles $30^\circ l$, $l=-2,\ldots,2$.
The predictions are reprojected into ERP coordinates.

For an ERP pixel $p$ observed in perspective view $v$, its observation
reliability is defined as
\begin{equation}
w_v(p)=\mathbf{1}_{\mathrm{valid}(v,p)}
[\cos\alpha_v(p)]_+^\eta\,
\operatorname{clip}\!\left(\frac{d_v(p)}{\delta},0,1\right),
\label{eq:weight}
\end{equation}
where $w_v(p)$ is the reliability of $p$ in view $v$, and
$\mathbf{1}_{\mathrm{valid}(v,p)}$ is 1 for a valid projection and 0 otherwise.
The notation $[x]_+=\max(x,0)$ retains non-negative values. The angle
$\alpha_v(p)$ is measured from the
optical axis, with $\eta$ controlling angular decay; $d_v(p)$ is the nearest
boundary distance, normalized by $\delta$. Finally,
$\operatorname{clip}(x,0,1)=\min\{1,\max\{0,x\}\}$. Thus, high reliability
requires a valid observation near the optical axis and away from view
boundaries.

Let $y(p)$ be the ERP row of pixel $p$, $\phi(y)$ its latitude,
$a_y=\max\{\epsilon,\cos\phi(y)\}$ the spherical-area weight of row $y$, and
$\mu(A)=\sum_{p\in A}a_{y(p)}$ the weighted area of region $A$. For a mask
$M$ reprojected from view $v$ and a same-class candidate $N$, we compute
\begin{equation}
\begin{aligned}
\operatorname{IoU}_{\mathrm{sph}}(M,N)
&=\frac{\mu(M\cap N)}{\mu(M\cup N)},\\
\bar{\mathbf{u}}_M
&=\frac{\sum_{p\in M}w_v(p)\mathbf{u}(p)}
{\left\|\sum_{p\in M}w_v(p)\mathbf{u}(p)\right\|_2},
\end{aligned}
\label{eq:association}
\end{equation}
where $\mathbf{u}(p)$ is the unit spherical direction of $p$,
$\|\cdot\|_2$ is the Euclidean norm, and $\bar{\mathbf{u}}_M$ is the
normalized spherical center of $M$. Candidates are matched when
$\operatorname{IoU}_{\mathrm{sph}}(M,N)\geq\tau_h$, or when
$\operatorname{IoU}_{\mathrm{sph}}(M,N)\geq\tau_m$ and
$\arccos(\bar{\mathbf{u}}_M^\top\bar{\mathbf{u}}_N)\leq\tau_\theta$.
Here, $\tau_h$, $\tau_m$, and $\tau_\theta$ are the high-overlap,
moderate-overlap, and angular-distance thresholds, respectively.

After association, global instance $g$ receives a pixel-wise support score
\begin{equation}
S_g(p)=\sum_{v\in\mathcal{O}_g}
w_v(p)\mathbf{1}[p\in M_{v,g}],
\label{eq:support}
\end{equation}
where $\mathcal{O}_g$ is the set of source-view observations, $M_{v,g}$ is
the candidate mask assigned to $g$ in view $v$, and
$\mathbf{1}[p\in M_{v,g}]$ equals 1
when $p$ belongs to that mask and 0 otherwise. Thing conflicts are resolved
by maximum support, after which category-wise stuff predictions fill
unassigned regions.

\subsection{Spherical Geometry-Aware Relation Reasoning}
\label{ssec:spherical_relation}

SGRR builds on the mask-prompted transformer encoder
in~\cite{Lorenz2024DSFormer}. A visual backbone and feature pyramid produce
an ERP feature map partitioned into patches $P_j$, whose projected embeddings
are $\mathbf{f}_j$. For ordered pair $(s,o)$, the masks, boxes, and object
labels condition the patch sequence, and the final classification token gives
the contextual pair feature $\mathbf{h}_{s,o}$.

For node $i\in\{s,o\}$, its spherical mass in patch $P_j$ is
$\mu_{i,j}=\sum_{p\in P_j}M_i(p)a_{y(p)}$. We obtain
\begin{equation}
\begin{aligned}
\mathbf{z}_i
&=\sum_j\frac{\mu_{i,j}}{\sum_r\mu_{i,r}}\mathbf{f}_j,\\
\widetilde{\mathbf{h}}_{s,o}
&=\mathbf{h}_{s,o}+
\operatorname{MLP}([\mathbf{z}_s;\mathbf{z}_o]),
\end{aligned}
\label{eq:region}
\end{equation}
where $j$ and $r$ index feature patches, $\mathbf{z}_i$ is the
spherical-area-weighted node feature, $[\,;\,]$ denotes feature
concatenation, and $\operatorname{MLP}$ is a multilayer perceptron. The output
$\widetilde{\mathbf{h}}_{s,o}$ is the augmented subject--object pair
representation.

For sampled ordered pairs $(s,o)\in\mathcal{B}$,
$\mathbf{y}_{s,o}\in\{0,1\}^{C}$ is the multi-hot foreground-predicate target,
and $y_{s,o}^{\emptyset}=\mathbf{1}[\sum_c y_{s,o,c}=0]$ indicates no
relation. Let $p_{s,o,c}$ and $p_{s,o}^{\emptyset}$ be the corresponding
predicted probabilities. Using binary cross-entropy
$\ell_{\mathrm{BCE}}(y,p)=-y\log p-(1-y)\log(1-p)$, we optimize
\begin{equation}
\begin{aligned}
\mathcal{L}_{\mathrm{rel}}
=\frac{1}{|\mathcal{B}|}
\sum_{(s,o)\in\mathcal{B}}\bigg[
&\sum_{c=1}^{C}
\ell_{\mathrm{BCE}}(y_{s,o,c},p_{s,o,c})\\
&+\lambda_{\emptyset}
\ell_{\mathrm{BCE}}(y_{s,o}^{\emptyset},p_{s,o}^{\emptyset})\bigg],
\end{aligned}
\label{eq:relation_loss}
\end{equation}
where $C$ is the number of foreground predicates and $\lambda_{\emptyset}$
weights no-relation supervision. For image $n$, $\mathcal{P}_n^+$ and
$\mathcal{N}_n$ denote positive and non-self negative pairs. All positives
are retained, negatives are sampled uniformly without replacement at ratio
$\rho$, and $\mathcal{B}$ is their union over the mini-batch. Multiple
predicates on one pair share a multi-hot target.

At inference, shifting the ERP image and masks by $W/2$ forms a second
longitude branch. Superscripts $(0)$ and $(\pi)$ denote the original and
shifted branches, which share parameters. We compute
\begin{equation}
\begin{aligned}
\bar{p}_{s,o}^{\emptyset}
&=\tfrac{1}{2}\left(
p_{s,o}^{\emptyset,(0)}+p_{s,o}^{\emptyset,(\pi)}
\right),\\
q_{s,o,c}
&=\left(1-\bar{p}_{s,o}^{\emptyset}\right)p_{s,o,c}^{(0)},
\end{aligned}
\label{eq:dual}
\end{equation}
where $\bar{p}_{s,o}^{\emptyset}$ is the fused no-relation probability and
$q_{s,o,c}$ is the final score used to rank predicate $c$ for pair $(s,o)$.
Thus, the two branches fuse relation-existence evidence while retaining the
original branch's foreground-predicate distribution.

\section{Experiments}
\label{sec:experiments}

\subsection{PanoPSG Dataset and Experimental Setup}
\label{ssec:dataset}

We introduce PanoPSG, a high-resolution panoramic panoptic scene graph
dataset comprising 340 ERP panoramas, each with full $360^\circ$ coverage
and a resolution of $11968\times5984$ pixels (Table~\ref{tab:dataset}).
Reviewer-verified masks were delineated in CVAT, while a dedicated platform
supported high-resolution relation annotation, persistent node identities,
and multiple predicates per ordered pair. A second review checked endpoint
validity, duplicate triplets, and mask--relation alignment.

\begin{table}[!t]
\centering
\caption{Statistics of the PanoPSG dataset.}
\label{tab:dataset}
\tablebodyfont
\setlength{\tabcolsep}{3pt}
\begin{tabular*}{\columnwidth}{@{\extracolsep{\fill}}lr@{}}
\toprule
Property & Statistics \\
\midrule
Panoramas (train / val) & 340 (272 / 68) \\
Resolution & $11968\times5984$ \\
Pixels / panorama & 71.62 M \\
Node classes (predefined / observed) & 133 / 82 \\
Nodes (thing / stuff) & 10,571 (4,043 / 6,528) \\
Relations / predicates & 6,632 / 21 \\
Mean nodes / relations per panorama & 31.09 / 19.51 \\
\bottomrule
\end{tabular*}
\end{table}

Each panorama contains an average of 31.09 nodes and 19.51 directed
relations, respectively $2.83\times$ and $3.48\times$ the PSG averages of
11.0 and 5.6~\cite{Yang2022PSG}. Together with 71.62 million pixels per image
and full-scene coverage, this density supports evaluating panoramic panoptic
segmentation and relation reasoning in complex scenes.For implementation, we set $\epsilon=10^{-6}$, $\eta=2$, $\delta=64$,
$\lambda_{\emptyset}=4$, and $\rho=1$.

SGPS is evaluated on all 340 panoramas using
panoptic quality (PQ)~\cite{Kirillov2019Panoptic}, spherical-area-weighted
mean intersection over union (Sphere-mIoU), seam-region mIoU (Seam-mIoU),
instance recall at an IoU threshold of 0.5 (IR@0.5), and mean instance recall
(mIR@0.5). Sphere-mIoU accumulates an $a_y$-weighted confusion matrix over
the dataset and averages class IoU over ground-truth-present classes.
Seam-mIoU is computed within the two $5^\circ$ strips adjacent to the ERP
seam. IR@0.5 uses one-to-one, same-category matching and excludes crowd thing
instances. IR is micro-averaged over eligible instances, whereas mIR is
macro-averaged over ground-truth-present thing categories.

Relation models use an image-disjoint split of 272 training and 68 validation
panoramas, a 30-epoch training budget, and best-validation checkpoint
selection. Predicate classification (PredCls) uses ground-truth masks and
object labels, isolating relation reasoning from segmentation errors. Although
the annotations permit multiple predicates for each ordered pair,
graph-constrained evaluation retains only
the highest-scoring predicate~\cite{Yang2022PSG,Lorenz2024DSFormer}. Recall
at $K$ (R@$K$) measures the recovery of ground-truth triplets among the
top-$K$ predictions per image. Mean recall (mR@$K$) macro-averages
predicate-specific recall over the 21 predicates, providing a
predicate-balanced complement to aggregate retrieval
\cite{Lorenz2023Haystack,Lorenz2024DSFormer}. We report
$K\in\{20,50\}$ over 1,325 validation relations. Under scene graph detection
(SGDet), node matching requires identical object labels and original-resolution
mask IoU $>0.5$. Unmatched predictions remain in the top-$K$ ranking.
Together, PredCls and SGDet distinguish relation-reasoning gains from
improvements inherited from panoramic node construction while retaining the
same recall-based evaluation.

\subsection{Spherical Geometry-Aware Panoptic Segmentation}
\label{ssec:node_results}

Table~\ref{tab:segmentation} compares first-stage panoptic segmentation
methods on PanoPSG. Direct ERP, Cubemap-6, Basic Fusion, and our method use
the same frozen MaskDINO segmentation model. Basic Fusion applies basic
fusion to the best-performing single-FoV configuration.

\begin{table}[!t]
\centering
\caption{Panoptic segmentation performance on PanoPSG (\%).}
\label{tab:segmentation}
\tablebodyfont
\setlength{\tabcolsep}{2.2pt}
\begin{tabular*}{\columnwidth}{@{\extracolsep{\fill}}lccccc@{}}
\toprule
Method &
PQ $\uparrow$ &
\shortstack{Sphere-\\mIoU} $\uparrow$ &
\shortstack{Seam-\\mIoU} $\uparrow$ &
IR@0.5 $\uparrow$ &
mIR@0.5 $\uparrow$ \\
\midrule
Direct ERP & 25.02 & 34.81 & 30.82 & 40.26 & 19.67 \\
Cubemap-6 & 33.22 & 45.47 & 47.71 & 61.92 & 43.29 \\
Basic Fusion & 28.89 & 45.97 & 48.81 & 72.71 & 58.67 \\
\textbf{Ours} & \textbf{39.30} & \textbf{56.04} & \textbf{61.11}
& \textbf{74.74} & \textbf{64.65} \\
\bottomrule
\end{tabular*}
\end{table}

Our method leads all five metrics in Table~\ref{tab:segmentation}. Compared
with Basic Fusion under the same view configuration, it improves PQ by 10.41
percentage points, corresponding to a relative improvement of 36.03\%. It
also gains 10.07 points in Sphere-mIoU, 12.30 points in Seam-mIoU, and 5.98
points in mIR@0.5. With the segmentation model and view sampling held fixed,
these results demonstrate the contribution of reliability-weighted spherical
association and fusion to panoptic quality, spherical semantic accuracy, and
seam continuity.

\subsection{Spherical Geometry-Aware Relation Reasoning}
\label{ssec:predcls}

Table~\ref{tab:predcls_comparison} compares the proposed relation reasoning
model with IMP~\cite{Xu2017IMP}, MOTIFS~\cite{Zellers2018Motifs},
VCTree~\cite{Tang2019VCTree}, and DSFormer~\cite{Lorenz2024DSFormer} under
the shared PredCls protocol.

\begin{table}[!t]
\centering
\caption{Relation-model comparison under PredCls on PanoPSG (\%).}
\label{tab:predcls_comparison}
\tablebodyfont
\setlength{\tabcolsep}{4.0pt}
\begin{tabular*}{\columnwidth}{@{\extracolsep{\fill}}lcccc@{}}
\toprule
Model &
R@20 $\uparrow$ &
R@50 $\uparrow$ &
mR@20 $\uparrow$ &
mR@50 $\uparrow$ \\
\midrule
IMP~\cite{Xu2017IMP} & 33.77 & 47.16 & 14.45 & 21.13 \\
MOTIFS~\cite{Zellers2018Motifs} & 49.92 & 63.59 & 28.40 & 36.42 \\
VCTree~\cite{Tang2019VCTree} & \textbf{53.03} & 67.85 & 30.91 & 40.15 \\
DSFormer~\cite{Lorenz2024DSFormer} & 45.06 & 68.60 & 25.50 & 53.49 \\
\textbf{Ours} & 48.68 & \textbf{70.57} & \textbf{35.18}
& \textbf{59.48} \\
\bottomrule
\end{tabular*}
\end{table}

Our method achieves the highest R@50 of 70.57\%, mR@20 of 35.18\%, and
mR@50 of 59.48\%. Relative to DSFormer, mR@20 and mR@50 increase by 9.68
and 5.99 percentage points, respectively, with the latter corresponding to an
11.20\% relative improvement. VCTree achieves the highest R@20, whereas our
method provides the strongest results for R@50 and both predicate-balanced
metrics. These results demonstrate that incorporating spherical geometry into
the relation pathway improves both aggregate and predicate-balanced relation
retrieval.

Table~\ref{tab:e2e} reports the SGDet ablation, where both panoptic segments
and directed relations are predicted.

\begin{table}[!t]
\centering
\caption{Component ablation under SGDet on PanoPSG (\%).}
\label{tab:e2e}
\tablebodyfont
\setlength{\tabcolsep}{3.0pt}
\begin{tabular*}{\columnwidth}{@{\extracolsep{\fill}}cccccc@{}}
\toprule
SGPS & SGRR &
R@20 $\uparrow$ &
R@50 $\uparrow$ &
mR@20 $\uparrow$ &
mR@50 $\uparrow$ \\
\midrule
$\times$ & $\times$
& 25.43 & 29.28 & 15.01 & 22.25 \\
$\checkmark$ & $\times$
& 27.02 & 42.79 & 14.62 & 27.00 \\
$\checkmark$ & $\checkmark$
& \textbf{29.13}
& \textbf{45.13}
& \textbf{18.45}
& \textbf{34.04} \\
\bottomrule
\end{tabular*}
\end{table}
Replacing Direct ERP segmentation with SGPS while retaining the baseline
relation model increases R@50 from 29.28\% to 42.79\% and mR@50 from 22.25\%
to 27.00\%. Adding SGRR further raises these metrics to 45.13\% and 34.04\%,
respectively. The complete framework therefore improves R@50 by 15.85 points
and mR@50 by 11.79 points over the baseline; the latter corresponds to a
52.99\% relative improvement. These results confirm complementary gains from
SGPS and SGRR under SGDet.
\section{Conclusion}
\label{sec:conclusion}

We presented a two-stage spherical geometry-aware panoramic PSGG framework
combining SGPS and SGRR, and introduced PanoPSG with 340 panoramas, 10,571
nodes, and 6,632 directed relations. Against corresponding baselines, SGPS
improves PQ by 36.03\% under matched views, SGRR improves mR@50 by 11.20\%
under PredCls, and the full framework improves mR@50 by 52.99\% under SGDet.
SGPS improves graph-node consistency, while SGRR reduces sensitivity to
latitude-dependent area and seam placement. These complementary gains
validate spherical geometry for panoramic PSGG. The framework provides a
unified basis for geometry-consistent scene graphs from high-resolution
panoramas.

\clearpage
\bibliographystyle{IEEEbib}
\bibliography{strings,refs}

\end{document}